\documentclass[sigconf]{acmart}
\AtBeginDocument{%
  }

\copyrightyear{2025}
\acmYear{2025}
\setcopyright{acmlicensed}\acmConference[WWW Companion '25]{Companion Proceedings of the ACM Web Conference 2025}{April 28-May 2, 2025}{Sydney, NSW, Australia}
\acmBooktitle{Companion Proceedings of the ACM Web Conference 2025 (WWW Companion '25), April 28-May 2, 2025, Sydney, NSW, Australia}
\acmDOI{10.1145/3701716.3718375}
\acmISBN{979-8-4007-1331-6/2025/04}

\usepackage{graphicx}
\usepackage{xcolor} 
\definecolor{LightGray}{gray}{0.9}
\usepackage{makecell}
\usepackage{siunitx}

\begin{document}

\title["Only ChatGPT gets me": An Empirical Analysis]{"Only ChatGPT gets me": An Empirical Analysis of GPT versus other Large Language Models for Emotion Detection in Text}

\author{Florian Lecourt}
\affiliation{%
  \institution{LIRMM UM5506 - CNRS, Université de Montpellier}
  \city{Montpellier}
  \country{France}
}

\author{Madalina Croitoru}
\affiliation{%
  \institution{LIRMM UM5506 - CNRS, Université de Montpellier}
  \city{Montpellier}
  \country{France}
}

\author{Konstantin Todorov}
\affiliation{%
  \institution{LIRMM UM5506 - CNRS, Université de Montpellier}
  \city{Montpellier}
  \country{France}
}

\renewcommand{\shortauthors}{Florian Lecourt, Madalina Croitoru, and Konstantin Todorov}

\begin{abstract}
  This work investigates the capabilities of large language models (LLMs) in detecting and understanding human emotions through text. Drawing upon emotion models from psychology, we adopt an interdisciplinary perspective that integrates computational and affective sciences insights. The main goal is to assess how accurately they can identify emotions expressed in textual interactions and compare different models on this specific task. This research contributes to broader efforts to enhance human-computer interaction, making artificial intelligence technologies more responsive and sensitive to users' emotional nuances. By employing a methodology that involves comparisons with a state-of-the-art model on the GoEmotions dataset, we aim to gauge LLMs' effectiveness as a system for emotional analysis, paving the way for potential applications in various fields that require a nuanced understanding of human language.
\end{abstract}



\maketitle


\section{Introduction}

The advent of artificial intelligence technologies, in particular conversational agents such as ChatGPT, has profoundly transformed the way we interact with machines \cite{obrenovic_generative_HCI_2024}. These agents, designed to simulate human conversations, now play a crucial role in various fields, from customer service \cite{jonnala2024large} to personal assistance \cite{xu2023can}. These novel technologies come with novel challenges for the AI community, among which we focus on one in particular: the ability to capture and correlate emotional expressions by machines and the ability of machines to express emotions and empathic behavior themselves \cite{ferscha2016research}. 

This work aims to provide a rigorous and detailed assessment of several LLMs, including GPT and LLama, and emerging models such as Gemini, Mistral, and Phi-3, focusing on their ability to detect and respond to emotions. Given that ChatGPT became "the fastest-growing app of all time" \cite{rudolph2023war}, we place particular emphasis on the GPT architecture it is based on. By cross-referencing the results of different evaluation methods, we aim to identify avenues of improvement to make conversational agents more empathetic and better adapted to users' needs. Our methodological approach thus aims to answer the question \textit{"How effectively do various large language models detect and classify human emotions from text compared to a state-of-the-art emotion detection model, using macro F1 score as an objective metric?"}. In contexts such as mental health, customer support, and social interactions, empathy, and emotional understanding are essential \cite{shao2023empathetic}.

The paper is structured as follows. Section \ref{emotionModels} introduces the psychological emotion models that form the conceptual foundation for our analysis. Section \ref{emotionDataset} presents the emotion datasets used for training and evaluating AI models. Section \ref{transformers} provides a quick overview of transformer-based architectures, including GPT, BERT, and other LLMs. Section \ref{evaluation} details our evaluation methodology, experiments, and results, covering prompt engineering techniques and cross-model comparisons. We conclude in Section \ref{conclusion} with comments on current findings and directions for future research.

\section{Emotion Models} \label{emotionModels}

In this section of the analysis, the term \textbf{model} refers to an emotion model, as understood in the field of psychology. We begin by disambiguating terminology to avoid confusion with a possible alternative meaning in Computer Science. According to Yadollahi et al. \cite{yadollahi_current_2018}, referencing the work of Fox \cite{Fox_Emotion_Science}, the terms emotion, mood, feeling, and affect are described in neuroscience as follows: 


\begin{itemize} 
\item  \textbf{Emotion}: A discrete and consistent response to internal or external events that have a particular significance for the organism; emotion has a short-term duration.
\item  \textbf{Mood}: a diffuse affective state that compared to emotion is usually less intense but with longer duration. 
\item  \textbf{Feeling}: A subjective representation of emotions, private to the individual experiencing them; similarly to emotion, it has a short-term duration.
\item  \textbf{Affect}: an encompassing term used to describe the topics of emotion, feelings, and moods together.
\end{itemize}

The terms \textbf{Emotion} and \textbf{Affect} are the most important here, as their uses will be found in the following works. We will now turn our attention to the various emotion models from the field of psychology. In their review of emotion models, Sreeja and Mahalakshmi \cite{p_s_emotion_2017} distinguish two categories of models:

\begin{itemize} 
\item  \textbf{Categorical (also called Discrete) :} These models feature several distinct emotions.

\item  \textbf{Dimensional:} These models represent emotions on continuous dimensions rather than discrete states.
\end{itemize}

According to Yadollahi et al., "while psychologists do not agree on what model describes more accurately the set of basic emotions, the model suggested by Ekman et al., with six emotions, is the most widely used in computer science research" \cite{yadollahi_current_2018}. For Paul Ekman, this model identifies six basic emotions that are universal and recognizable by all human cultures: joy, sadness, anger, fear, surprise, and disgust \cite{ekman_argument_1992}. Ekman developed this model from his research into facial expressions and human emotions. His first study in this domain was in 1970, where Ekman asked New Guineans to associate photographs and emotions \cite{ekman1970universal}. The study's sample is of 189 adults and 130 children. Following the study's protocol, the experiment showed three photographs to a test subject, told a story concerning one of the emotions in Ekman's taxonomy, and then asked the subject to pick the photograph that fits the story. Ekman states, "The results were very clear, supporting our hypothesis that there is a pan-cultural element in facial expressions of emotion."

Before Ekman, Tomkins proposed a model comprising eight fundamental affects, identified by different facial expressions: Interest-Excitement, Pleasure-Joy, Surprise, Distress-Anguish, Fear-Terror, Shame-Humiliation, Contempt-Disgust and Anger-Rage \cite{tomkins_affect_1962}. For Tomkins, emotions "consist of one or more affects in combination with cognitive or drive states in a manner that colors, flavors, or inflects the affects" \cite{frank2020silvan}, corresponding to the definition we gave to the term affect. In each pair, the first term corresponds to "the most characteristic description as experienced at low [...] intensity", and the second term to the one experienced at high intensity. Tomkins used compound names for these affects to describe the expressed affect as characteristic as possible.

Building on Tomkins' work, Lövheim has developed a dimensional model represented by a cubic structure \cite{lovheim_new_2011}. Each corner of this cube corresponds to an affect described by Tomkins. In this representation, each emotion is positioned along orthogonal axes defined by the levels of three monoamines: dopamine (DA), serotonin (5-HT), and noradrenaline (NE). For example, the Anger-Rage affect is characterized by high levels of dopamine and noradrenaline but low serotonin levels. According to Lövheim, the advantage of this dimensional model lies in its ability to correlate directly with the field of neurobiology.

\begin{figure}[H]
    \centering
    \includegraphics[width=1\linewidth]{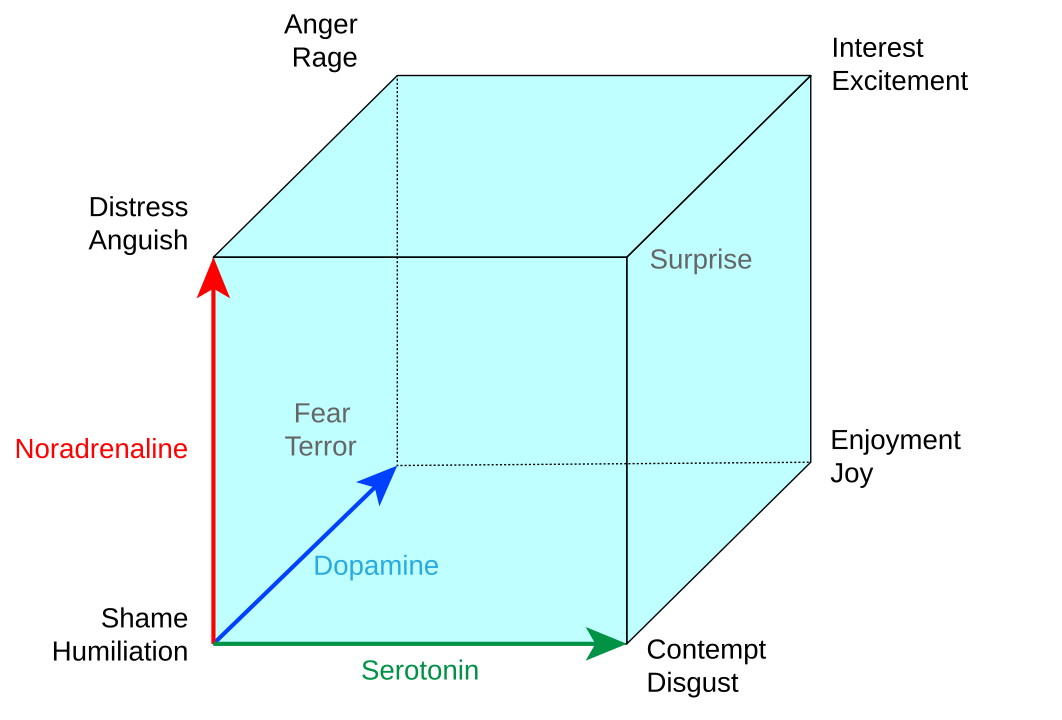}
    \caption{Graphical representation of the Lövheim model \cite{LovheimCubeEmotions2024}.}.
    \label{fig:cube-lovheim}
\end{figure}

\vspace{-1.cm} 

Lövheim compares his approach with Plutchik's dimensional model in the introductory article to his model. Plutchik describes an eight-emotion model: Fear, Anger, Joy, Sadness, Acceptance/Trust, Disgust, Anticipation, and Surprise \cite{plutchik_psychoevolutionary_1982}. He justifies this choice by linking each emotion to a biological factor. In 1991, Plutchik describes an experiment in which 30 university students rate the intensity of different emotions on a scale from 1 to 11 \cite{plutchik_emotions_1991}. The list includes the eight primary emotions and their synonyms. Based on the data collected, he proposes a model in which the most intense emotions are represented closer to the center and with more saturated colors than those of less intense emotions. Plutchik points out that the opposing primary emotions in this emotional wheel are complementary and that their combination produces a neutral psychic or biological state comparable to gray. The 3D version of Plutchik's model, which represents intensity on the depth axis, illustrates these concepts more explicitly.

\begin{figure}[H]
    \centering
    \includegraphics[width=0.6\linewidth]{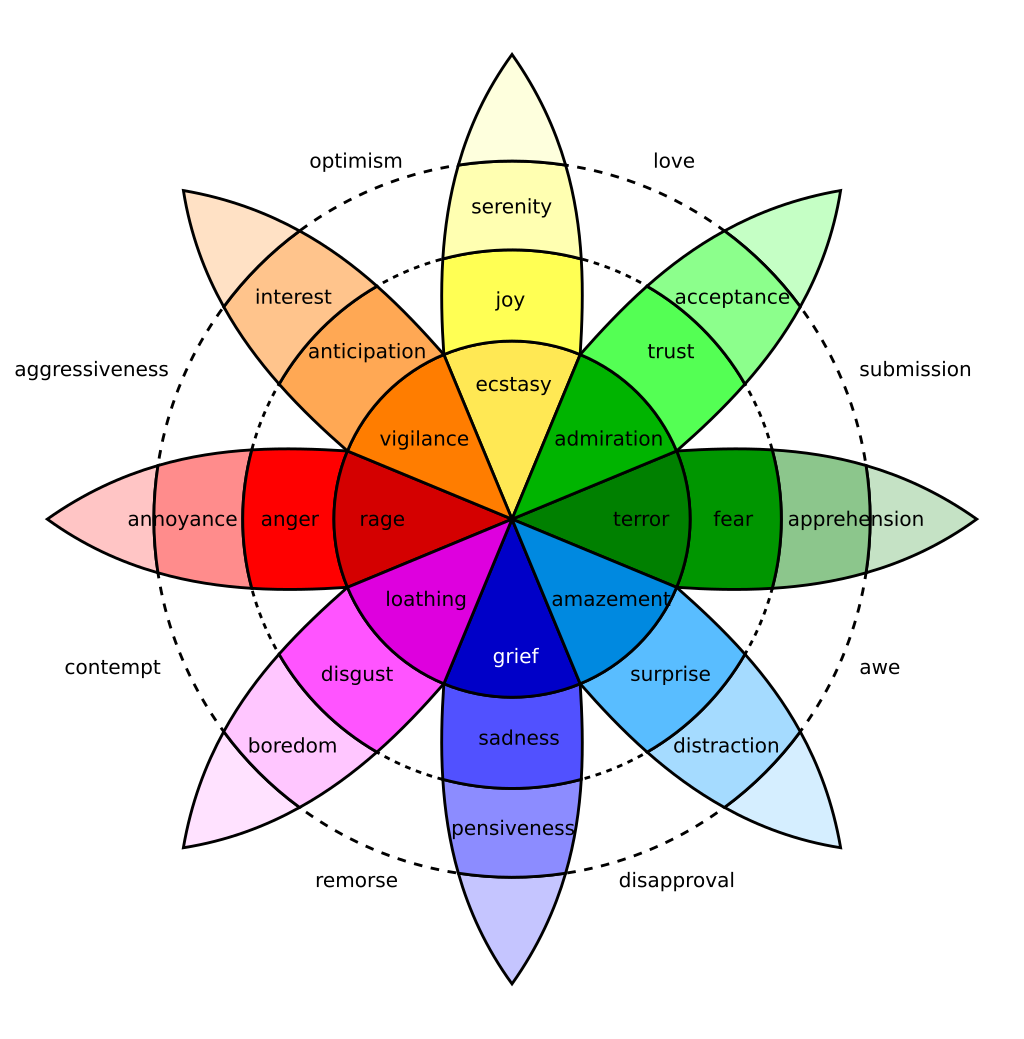}
    \caption{2D representation of the Plutchik model \cite{Plutchik_wheel_2024}}
    \label{fig:plutchik-wheel}
\end{figure}

\begin{figure}[H]
    \centering
    \includegraphics[width=0.5\linewidth]{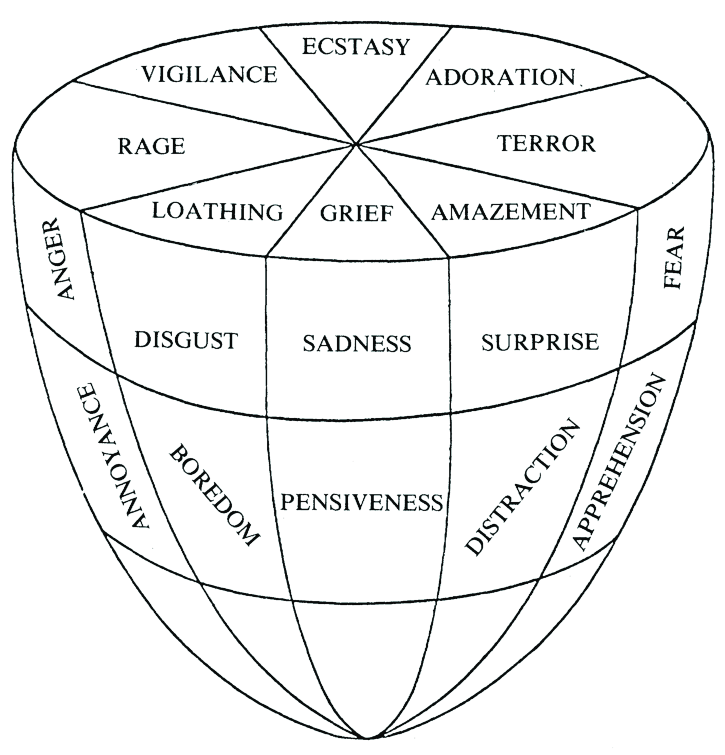}
    \caption{3D representation of the Plutchik model \cite{article}}
    \label{fig:pluchik_3d}
\end{figure}

\begin{table}[h!]
\setlength{\tabcolsep}{3pt}
\resizebox{\linewidth}{!}{%
\begin{tabular}{|| l |l|l|l|l ||} 
    \toprule
 & Ekman & Tomkins & Lövheim & Plutchik \\
    \hline
\midrule
    Joy & \checkmark & \checkmark & \checkmark & \checkmark \\
    Anger & \checkmark & \checkmark & \checkmark & \checkmark \\
    Fear & \checkmark & \checkmark & \checkmark & \checkmark \\
    Sadness & \checkmark & & & \checkmark \\
    Acceptance/Confidence & & & & \checkmark  \\
    Disgust & \checkmark & \checkmark & \checkmark & \checkmark \\
    Anticipation & & & & \checkmark  \\
    Surprise & \checkmark & \checkmark & \checkmark & \checkmark \\
    Distress & & \checkmark & \checkmark & \\
    Shame & & \checkmark & \checkmark & \\
    Interest & & \checkmark & \checkmark & \\
    Model type & Discrete & Discrete & Dimensional & Dimensional \\
\bottomrule
\end{tabular}%
}
\caption{Comparison of the emotions present in each model.}
\label{tab:Comparison_Models}
\end{table}

\vspace{-.5cm}
In Table \ref{tab:Comparison_Models}, we observe that the emotions common to the different models include Joy, Anger, Fear, Disgust, and Surprise. As mentioned in the introduction of this section, there is no consensus about which model best represents the spectrum of human emotions. When these models are used in computer science, the different emotions present in each model (or affects in the case of Tomkins and Lövheim) are used to create a taxonomy to annotate datasets. While categorical models are well-suited for such a purpose, information is inevitably lost during discretization in the case of dimensional models.

\section{Emotion Datasets} \label{emotionDataset}

Emotion detection uses specific datasets to train and evaluate emotional classification models efficiently and accurately. The GoEmotions dataset, developed by Google, is a collection of 58,000 Reddit comments, manually annotated to cover 27 emotional categories and one neutral category \cite{demszky_2020_goemotions}. This dataset stands out for its granularity, offering detailed and nuanced coverage of human emotions. Data were collected from 2005 to January 2019, excluding deleted and non-English comments. To limit bias, the data was partially filtered to reduce vulgarities while retaining those deemed essential for learning about negative emotions, limiting text length, and balancing the emotions represented. The final taxonomy of emotions was established through an iterative process to maximize the coverage of emotions expressed in the Reddit data while limiting the total number of emotions and their overlap. Initially, 56 emotional categories were considered. During iterative refinement, categories that the annotators rarely selected showed low concordance or were difficult to detect in the text were removed to improve clarity \cite{site_demszky_goemotions_2020}. Frequently suggested categories that were well represented in the data were added. This refinement process resulted in high annotation accuracy, with 94\% of examples having at least two annotators agreeing on at least one emotional label. As a result, GoEmotions includes 12 positive, 11 negative, and four ambivalent emotions, enabling GoEmotions to serve as a reliable resource for the fine classification of emotions in texts.


The CARER dataset is a less granular dataset than GoEmotions, featuring eight emotion labels (Joy, Surprise, Anticipation, Fear, Anger, Trust, Disgust, and Sadness) \cite{saravia-etal-2018-carer}. Unlike GoEmotions, each text, based on tweets, is associated with a unique emotion label. The eight labels used are the same as those described by Plutchik \cite{plutchik_psychoevolutionary_1982}. This feature is shared by the WRIME dataset \cite{kajiwara_wrime_2021}, composed of texts from various social networks, and GoodNewEveryone \cite{bostan_goodnewseveryone_2020}, which takes newspaper headlines and adds the labels Guilt, Love, Pessimism, Optimism, Pride and Shame, separating Surprise into Positive Surprise and Negative Surprise. For the latter, similarly to GoEmotions, newspaper headlines were annotated by comparing agreements between annotators.

The oldest and most cited dataset is the ISEAR dataset \cite{scherer_evidence_1994}. Based on Ekman's work and using the emotions described in it (replacing Surprise with Shame and Guilt), ISEAR is a dataset derived from psychological research to prove the universality and cultural variations of differential emotional response patterns. The various data come from a series of questionnaires taken in 37 different countries.

Whether for datasets based on the work of Plutchik and Ekman or for GoEmotions, the shared emotions are Joy, Anger, Fear, Sadness, and Disgust. Unlike the emotions shared by the different models, the emotion of Surprise is absent here due to its non-use in the ISEAR data, and the emotion of Sadness makes its appearance, already being a common emotion in the Ekman and Plutchik models.

\begin{table}[h!]
\setlength{\tabcolsep}{3pt}
\resizebox{\linewidth}{!}{%
\begin{tabular}{|| l |l|l|l|l|l ||} 
    \toprule
 & GoEmotions & CARER & WRIME & GoodNewsEveryone & ISEAR \\
    \hline
\midrule
    Admiration & \checkmark & & & & \\
    Amusement & \checkmark & & & & \\
    Anger & \checkmark & \checkmark & \checkmark & \checkmark & \checkmark \\
    Annoyance & \checkmark & & & \checkmark & \\
    Anticipation & & \checkmark & \checkmark & &  \\
    Approval & \checkmark & & & & \\
    Caring & \checkmark & & & & \\
    Confusion & \checkmark & & & & \\
    Curiosity & \checkmark & & & & \\
    Desire & \checkmark & & & & \\
    Disappoint-ment & \checkmark & & & & \\
    Disapproval & \checkmark & & & & \\
    Disgust & \checkmark & \checkmark & \checkmark & \checkmark & \checkmark \\
    Embarrass-ment & \checkmark & & & & \\
    Excitement & \checkmark & & & & \\
    Fear & \checkmark & \checkmark & \checkmark & \checkmark & \checkmark \\
    Gratitude & \checkmark & & & & \\
    Grief & \checkmark & & & & \\
    Guilt & & & & \checkmark & \checkmark \\
    Joy & \checkmark & \checkmark & \checkmark & \checkmark & \checkmark \\
    Love & \checkmark & & & \checkmark & \\
    Nervousness & \checkmark & & & & \\
    Neutral & \checkmark & & & & \\
    Optimism & \checkmark & & & \checkmark & \\
    Pessimism & & & & \checkmark & \\
    Pride & \checkmark & & & \checkmark & \\
    Realization & \checkmark & & & & \\
    Relief & \checkmark & & & & \\
    Remorse & \checkmark & & & & \\
    Sadness & \checkmark & \checkmark & \checkmark & \checkmark & \checkmark \\
    Shame & & & & \checkmark & \checkmark \\
    Surprise & \checkmark & \checkmark & \checkmark & \checkmark & \\
    Trust & & \checkmark & \checkmark & \checkmark & \\
\bottomrule
\end{tabular}%
}
\caption{Comparison of the emotions present in each dataset.}
\label{tab:Comparaison_Datasets}
\end{table}

\section{The Transformer models} \label{transformers}
Introduced by Vaswani et al. \cite{vaswani_2017_attention}, Transformer models have revolutionized NLP thanks to their innovative architecture, overcoming the limitations of previous approaches such as Recurrent Neural Networks (RNN) and Long Short-Term Memory (LSTM)  \cite{OverviewTransformerbasedModelsNLPTasks_gillioz_etal}. Those models are interesting notably due to their exceptional performance, which is now state-of-the-art in many fields, such as NLP \cite{TransformersEndHistoryNLP_chernyavskiy_etal} or audio processing \cite{koutini_efficient_2022}. Transformers are at the root of LLMs such as GPT and classifiers such as BERT. These families of models are used for emotion detection as well. 

\paragraph{}{Developed by OpenAI, GPT is an auto-regressive model \cite{radford2018improving}. This architecture generates text sequentially, predicting each subsequent word based on previously generated words. ChatGPT is a conversational agent, a chatbot, based on the GTP-3.5 model.}

\paragraph{}{In contrast, BERT is an example of an encoder model \cite{DBLP:journals/corr/abs-1810-04805}. Unlike GPT, BERT is specifically designed to understand and analyze language. It excels in text classification, comprehension, and sentiment analysis tasks.}

\subsection{Other LLMs}

In this work, we investigate the emotion detection capabilities of several LLMs. In addition to GPT, we examine Gemini, Gemma, LLaMA, Phi3, Mistral, and Mixtral. Here is a short description of those models:

\begin{itemize} 

\item \textbf{LLama \cite{touvron2023llama}:} LLama models are open-source LLMs distributed by Meta. They are designed to be computationally efficient and easy to fine-tune.

\item \textbf{Mistral \cite{jiang2023mistral}/ Mixtral \cite{jiang2024mixtral}:} Mistral and Mixtral are two LLM introduced by Mistral AI. Mistral outperforms LLama 2 on multiple benchmarks while maintaining faster inference. Mixtral is based on a sparse mixture-of-experts (SMoE). Each token is processed by two out of eight experts per layer, giving Mixtral effective access to large parameter spaces while only using 13B active parameters per inference step. Mixtral competes with GPT-3.5 on many benchmarks. 

\item \textbf{Gemma \cite{team2024gemma}/ Gemini \cite{team2024gemini}:} Developed by Google, Gemma, and Gemini represent two distinct approaches in the LLM ecosystem. Gemma models are open-source solutions designed for multilingual understanding and accessibility, making them adaptable to various applications. Gemini is a multimodal LLM crafted to excel at complex benchmarks, positioning itself as a strong competitor to high-performance models like Claude 3.0 or GPT-4.

\item \textbf{Phi-3 \cite{abdin2024phi}:} Phi-3, a kind of model brought by Microsoft, can be described as a Small Langue Model (SLM). Despite its relatively compact size, it is designed to achieve top-tier performance and rival larger models such as Mixtral and GPT-3.5. 

\end{itemize}

After establishing these models' foundational concepts and characteristics, we can now move on to evaluating their performances in emotion detection tasks.

\section{Evaluation and Results}\label{evaluation}

\subsection{Emotion Detection}

Natural Language Processing (NLP) is an essential branch of artificial intelligence devoted to understanding and manipulating human language by machines. NLP problems can be divided into two main categories: symbolic and statistical \cite{wermter_connectionist_1996}. Statistical approaches are the basis behind Transformer models and LLMs, so we focus on these methods here.

While traditional opinion mining, or sentiment analysis, classifies opinions as positive, negative, or neutral, emotion detection (ED) offers a more nuanced understanding of affective states \cite{bouazizi_sentiment_2016}. By moving beyond a binary or ternary scale, ED captures subtle emotional cues, paving the way for more empathetic and contextually aware AI applications.

\subsection{Chat-GPT and Emotion Detection}
After exploring Transformers models, BERT, LLMs, and emotion datasets such as GoEmotions, it is pertinent to look at the comparative evaluation of these models in the specific domain of emotion detection. The article \emph{ChatGPT: Jack of all trades, master of none} evaluates ChatGPT's performance on various NLP tasks, including emotion detection \cite{kocon_chatgpt_2023}. This evaluation compares ChatGPT with models considered to be state-of-the-art (SOTA) for the same tasks.

In the field of emotion detection, ChatGPT is evaluated as a classifier. Its performance is measured using the GoEmotions dataset. Given the variability in the numbers of each emotional class in this dataset, the F1 macro score is used as the evaluation metric. The F1 macro score is calculated as the arithmetic mean of the individual F1 scores for each class, where each F1 score is itself the harmonic mean of precision and recall for that class. This method enables a balanced evaluation by not favoring any particular class, regardless of their prevalence in the dataset. This property is essential in contexts where classes are unequally represented, as it prevents the bias towards majority classes that could distort the overall assessment of model performance. By balancing the influence of each class, the F1 macro encourages the development of models that effectively recognize all emotions, including less frequent ones, thus contributing to a richer understanding of the emotional nuances captured in the text.

\subsection{Reproduction of Results}
In the following section, we specifically seek to reproduce the results observed in Kocon's study \cite{kocon_chatgpt_2023} to verify the consistency of ChatGPT's performance in emotion detection, as described in this publication.

\begin{figure}[H]
    \centering
    \includegraphics[width=0.75\linewidth]{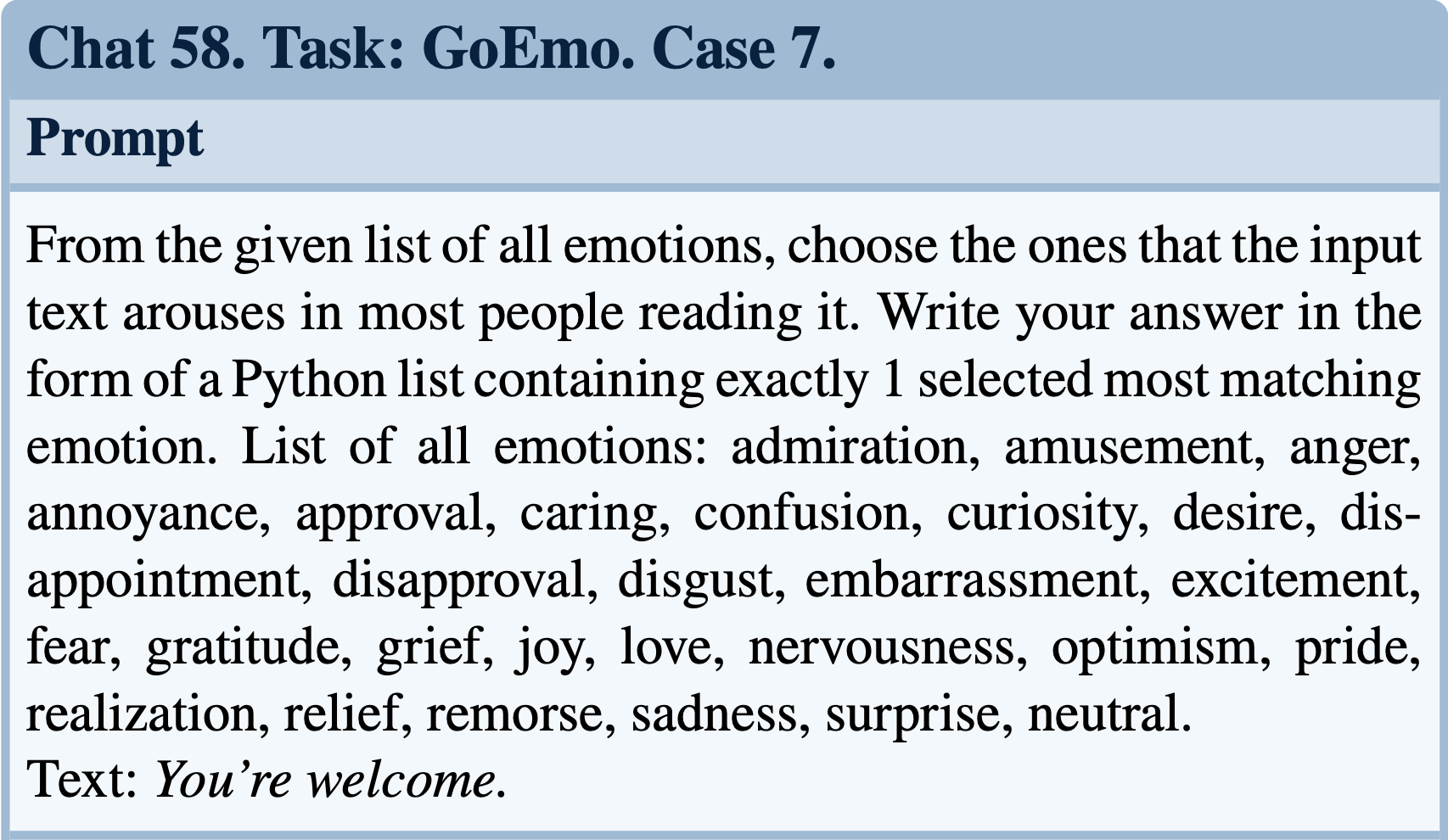}
    \caption{Example prompt \cite{kocon_chatgpt_2023}}
    \label{fig:prompt-sota-gpt}
\end{figure}

Firstly, the BERT model, referred to as SOTA, used by the article's authors, is tested to confirm its F1 macro score \cite{noauthor_monologgbert-base-cased-goemotions-original_nodate}. The second step is to use the OpenAI API to interact with GPT-3.5-Turbo, which is identical to the one on which ChatGPT is based. A specific prompt is sent through the API to evaluate ChatGPT, which then generates the model response. The structure of this prompt is inspired by the article, as illustrated in Figure \ref{fig:prompt-sota-gpt}. The response received from ChatGPT is then analyzed to calculate its F1 macro score. The BERT model and GPT-3.5-Turbo will be tested using the \textbf{test} set from the dataset GoEmotions.

Evaluated metrics include:
\begin{itemize}
    \item \textbf{ChatGPT macro F1 score (\%)}: Calculated as the average of the F1 scores for each class, this measures ChatGPT's overall performance across all classes regardless of their frequency of appearance. 
    
    \item \textbf{SOTA macro F1 score (\%)}: Measures the performance of the SOTA model for the same task. Calculated in the same way as ChatGPT's F1 macro.
    
    \item \textbf{Difference (pp)}: The difference in percentage points between the F1 macro scores of ChatGPT and the SOTA model.
    
    \item \textbf{Difficulty (\%)} : Defined as \[
    \text{Difficulty} = 100\% - F1_{\text{macro, SOTA}}
    \] This metric reflects the task's intrinsic difficulty based on the SOTA model's performance.
    
    \item \textbf{Loss (\%)} : Calculated as \[
    \text{Loss} = 100\% \times \frac{F1_{\text{macro, SOTA}} - F1_{\text{macro, ChatGPT}}}{F1_{\text{macro, SOTA}}}
    \] This metric shows the performance loss of ChatGPT compared with the SOTA model.
\end{itemize}

\begin{table*}[h!]
\setlength{\tabcolsep}{3pt}
\resizebox{\linewidth}{!}{%
\begin{tabular}{|| l | S|S|S|S|S ||} 
    \toprule
 & \textbf{ChatGPT macro F1 score (\%)} & \textbf{SOTA macro F1 score (\%)} & \textbf{Difference (pp)} & \textbf{Difficulty (\%)} & \textbf{Loss (\%)} \\
    \hline
\midrule
         Reference values & 25.55 & 52.75 & 27.20 & 47.25 & 51.56 \\
         Test1, Batch size 500 & 22.43 & 48.86 & 26.43 & 51.14 & 54.09 \\ 
         Test2, Batch size 1000 & 22.82 & 52.19 & 29.37 & 47.81 & 56.28 \\
         Test3, Batch size 2500 & 22.83 & 49.30 & 26.47 & 50.70 & 53.69 \\
         Test4, Entire dataset & 23.02 & 49.68 & 26.66 & 50.32 & 53.67 \\
\bottomrule
\end{tabular}
}
\caption{Comparison of Chat-GPT and SOTA model performance depending on the batch size.}
\label{tab:comparaison_resultats_1}
\end{table*}

Analysis of ChatGPT's performance in comparison with the SOTA model on the emotion detection task, as illustrated in Table \ref{tab:comparaison_resultats_1}, reveals a significant deviation from the performance of the SOTA model. This discrepancy is noticeable in all the contexts tested, with a performance loss of more than 50\% in all contexts. This observation suggests that, despite ChatGPT's advanced text generation capabilities, its performance in the specific emotion detection task remains substantially inferior to that of a model dedicated to this task, confirming the article's conclusions. The various tests were carried out with varying batch sizes due to the constraints imposed by the OpenAI API. In the following section, the batch size used for testing will be the one from \textbf{Test2}, as the results obtained for this test are the closest to the one in Kocon's paper.

\subsection{Evaluation Setting} 

The methodology of this study consists of several steps aimed at evaluating and improving the performance of ChatGPT for the emotion detection task. First, we thoroughly review prompt engineering techniques, building on approaches identified in the state of the art to optimize the instructions given to ChatGPT. The aim is to maximize its F1 macro score, a metric chosen to evaluate the model's accuracy in a balanced way across all emotional classes.

\begin{figure}[H]
\centering
\includegraphics[width=0.6\linewidth]{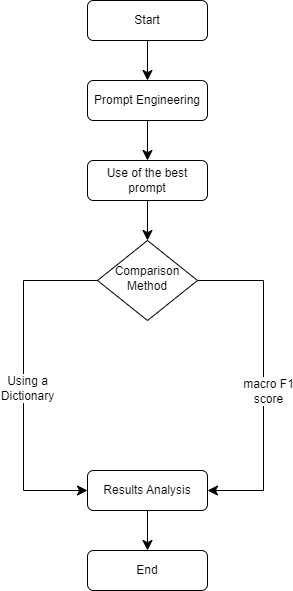}
\caption{Evaluation flowchart}
\label{fig:organization}
\end{figure}

Once the best prompt has been determined, we will compare ChatGPT's performance with other language models using the same optimized prompt. Once again, ChatGPT is represented by the GPT-3.5-turbo model, on which it is based. This comparison will enable us to situate ChatGPT in relation to other models in the specific context of emotion detection. Then, to check the results' robustness, we will employ complementary methods, such as integrating dictionaries to correct responses that do not appear in the list of 28 emotions. Figure \ref{fig:organization} shows a flowchart of the evaluation.

\subsection{Prompt Engineering}

\begin{figure}[h!]
    \centering
    \includegraphics[width=1\linewidth]{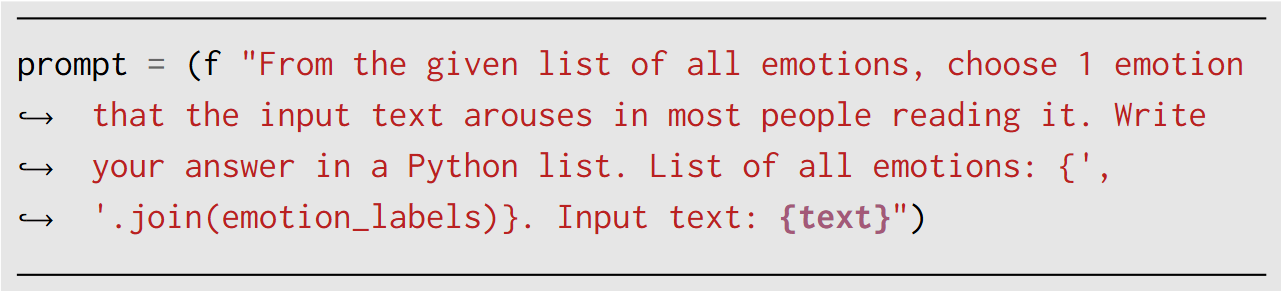}
    \caption{Original prompt}
    \label{fig:Prompt_0}
\end{figure}

To optimize GPT's performance in the emotion detection task, we explored several variants of prompts. Each variant aims to refine the instructions given to the model to improve the accuracy and consistency of responses. The four prompts used in this study are detailed below, each with specific adjustments to maximize the F1 macro score. The basic prompt (Figure \ref{fig:Prompt_0}) asks GPT to select a single emotion from a given list elicited by the text provided. This prompt serves as a starting point for evaluating the initial performance of the GPT model.

\begin{figure}[h!]
    \centering
    \includegraphics[width=1\linewidth]{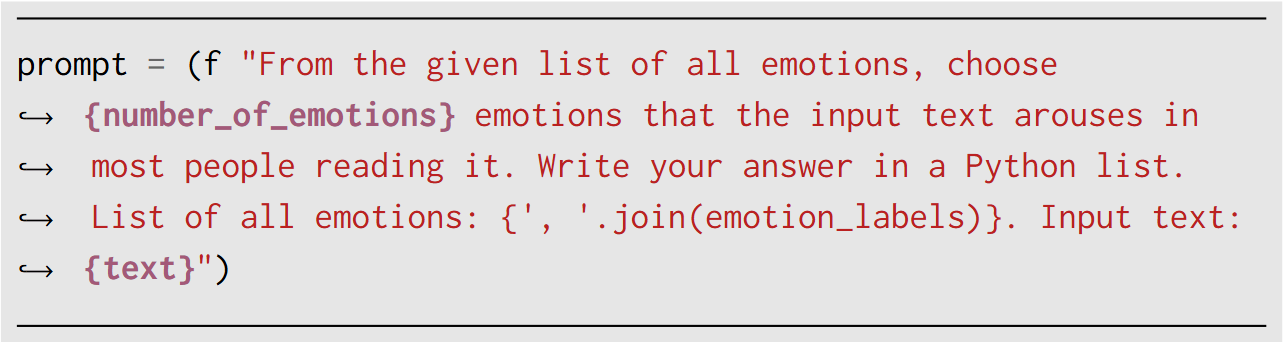}
    \caption{First Variant}
    \label{fig:Prompt_1}
\end{figure}

The new prompt, seen in Figure \ref{fig:Prompt_1},  adds a variable for the number of emotions to be identified, corresponding to the number of emotions annotated for the given text in the GoEmotions dataset. This approach better aligns GPT's responses with the dataset's annotations.

\begin{figure}[h!]
    \centering
    \includegraphics[width=1\linewidth]{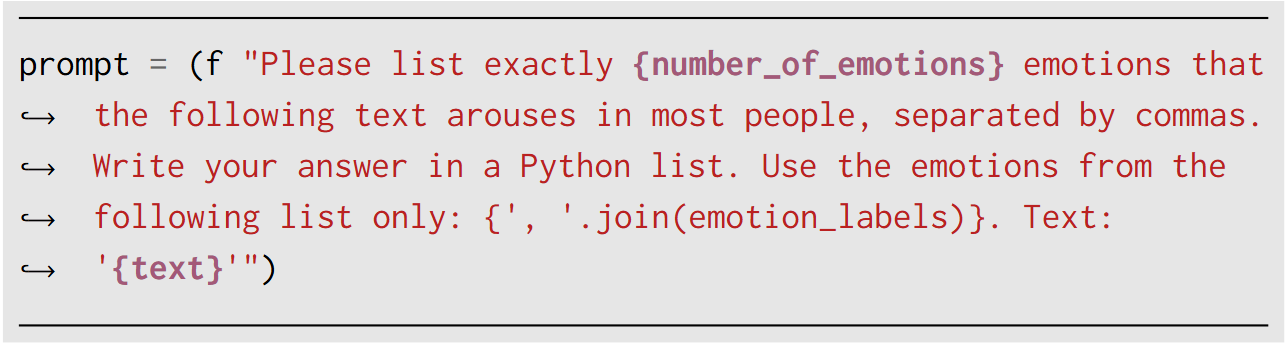}
    \caption{Second Variant}
    \label{fig:Prompt_2}
\end{figure}

For the next prompt, emphasis is placed on the exact number of emotions to be returned using the phrase "Please list exactly {number\_of\_emotions}." This formulation is intended to reduce ambiguity and encourage GPT to adhere strictly to the requested number of emotions. This prompt is illustrated in Figure \ref{fig:Prompt_2}.

\begin{figure}[h!]
    \centering
    \includegraphics[width=1\linewidth]{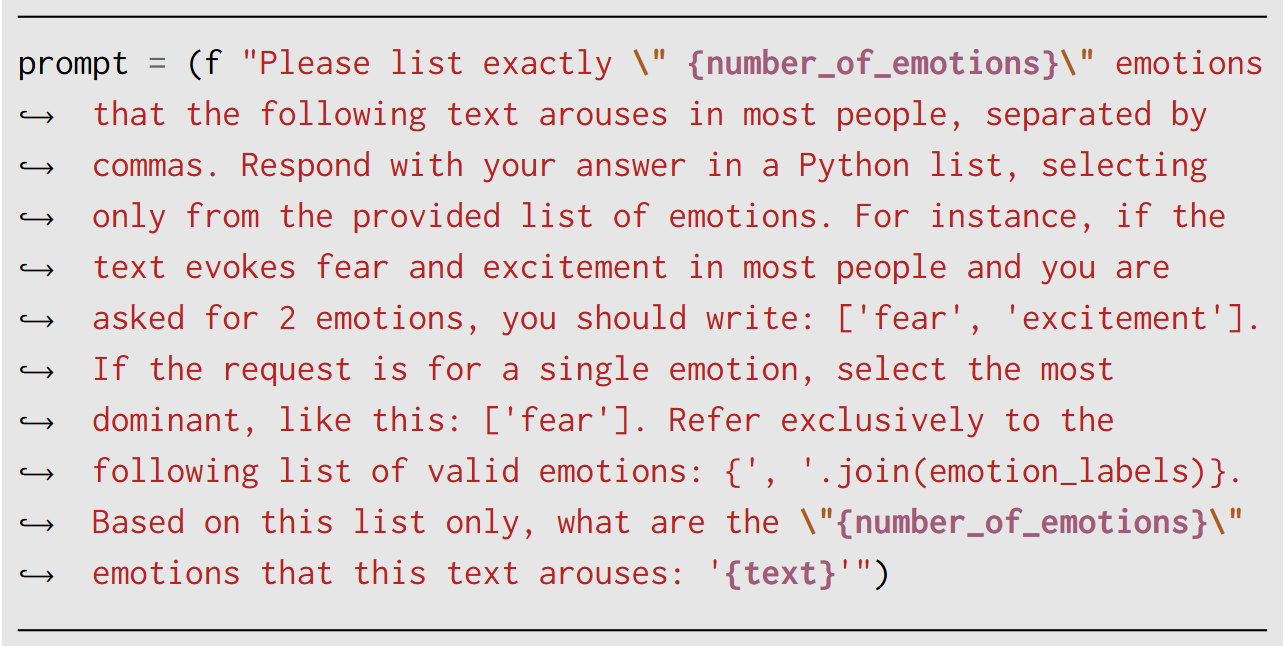}
    \caption{Third Variant}
    \label{fig:Prompt_3}
\end{figure}

The last prompt (Figure \ref{fig:Prompt_3}) repeats the methodology of the previous prompt while adding quotation marks around the number of emotions requested and providing an explicit example of the expected response format. This example is intended to clarify expectations further and guide GPT towards a correctly formatted response.

\begin{table}[h!]
\centering
\setlength{\tabcolsep}{3pt}
\resizebox{\linewidth}{!}{%
\begin{tabular}{|| l | S|S|S ||} 
    \toprule
     & \textbf{Model macro F1 score (\%)} & \textbf{Difference (pp)} & \textbf{Loss (\%)} \\
    \hline
\midrule
        Reference values & 22.82 & 29.37 & 56.28 \\
        Variant 1 & 27.28 & 24.91 & 47.73 \\
        Variant 2 & 26.14 & 26.05 & 49.91 \\
        Variant 3 & 28.97 & 23.22 & 44.49 \\
\bottomrule
\end{tabular}%
}
\caption{Comparison of Chat-GPT performance depending on the prompt used.}
\label{tab:Comparison_Prompts}
\end{table}

Table \ref{tab:Comparison_Prompts} shows that the last prompt achieves the highest F1 macro score. In the remainder of this study, we will use this prompt to explore Chat-GPT's performance in greater depth and compare it with other language models.

\subsection{Comparisons with other LLMs}
As mentioned in the previous subsection, we will now compare the F1 macro scores of Chat-GPT with those of other large language models (LLMs). The aim is to determine whether one model outperforms GPT-3.5-Turbo in the emotion detection task. The prompt in Figure \ref{fig:Prompt_3}, which gave the best results for Chat-GPT, will be used for these comparisons.

Gemini-1.5 results were obtained using a Google Colab provided by Google. The performance of Llama-3-70b and Mixtral-8x7b was measured via the Huggingchat API, as these models are too large to be run locally. The other results were obtained by running the models locally using the Ollama application and Python library.

\begin{table}[h!]
\setlength{\tabcolsep}{3pt}
\resizebox{\linewidth}{!}{%
\begin{tabular}{|| l | S|S|S ||} 
    \toprule
    Model name & \textbf{Model macro F1 score (\%)} & \textbf{Difference (pp)} & \textbf{Loss (\%)} \\
    \hline
\midrule
         GPT-3.5-Turbo &  28.97 & 23.22 & 44.49 \\
         GPT-4o &  30.95 & 21.24 & 40.70 \\
         Llama-2-7b &  20.24 & 31.95 & 61.22 \\
         Llama-3-8b &  20.60 & 31.59 & 60.53 \\
         Llama-3-70b &  27.20 & 24.99 & 47.88 \\
         Phi-3-4k &  25.23 & 26.96 & 51.66 \\
         Gemma-1.1-7b &  22.89 & 29.30 & 56.14 \\
         Gemma-2-9b &  24.21 & 27.98 & 53.61 \\
         Gemini-1.5 &  26.74 & 25.45 & 48.76 \\
         Mistral-7b &  25.14 & 27.05 & 51.83 \\
         Mixtral-8x7b & 23.82 & 28.37 & 54.36 \\
\bottomrule
\end{tabular}%
}
\caption{Comparing the performance of different language models.}
\label{tab:comparaison_resultats_2}
\end{table}

Analysis of the results presented in Table \ref{tab:comparaison_resultats_2} reveals significant differences between language model families. Models in the GPT family, including GPT-3.5-Turbo and GPT-4o, stand out for their overall superior performance in emotion detection. In particular, GPT-4o shows a slight improvement over GPT-3.5-Turbo, underlining the continued progress in this series.

The Llama family models, particularly Llama-3-70b, also show promising skills, albeit slightly inferior to those of the GPT models. Lighter versions, such as Llama-2-7b and Llama-3-8b, do not achieve the same level of performance, indicating a correlation between model size and emotion detection capabilities for this model family.

Google-developed models, such as Gemini-1.5, Gemma-1.1-7b, and Gemma-2-9b, show respectable results, although they do not surpass GPT models. However, this model family continues to offer a solid alternative with consistent performance.

The Mistral and Mixtral models show less competitive results compared to the GPT and Llama-3-70b models, although they have superior skills compared to the other Llama models.

Finally, the Phi-3 model, developed by Microsoft, shows competitive performance, positioning itself between the Llama and Google models regarding the macro F1 score. Phi-3 is a Small Language Model (SLM), a category of models developed by Microsoft to offer capabilities similar to those of large language models but with reduced size and resource requirements. SMLs thus offer an efficient alternative to LLMs for specific tasks.

In summary, GPT models dominate in terms of performance, followed by Llama and Google models. Though inferior performers, Mistral, Mixtral, and Phi-3 may offer viable alternatives.

Although the macro F1 score or Accuracy are widely used and enables a standardized performance comparison between different models, they have certain limitations when it comes to capturing the subtlety of the errors made by these models. In particular, they treat each error equally without considering the semantic proximity between predicted and true emotions. This binary approach to errors is problematic in emotion detection, where certain emotions are intrinsically closer to each other, especially in fine-granulated datasets such as GoEmotions.

\subsection{Using a Dictionary}

ChatGPT and the other LLMs sometimes respond outside the requested emotions list. In the previous results, these responses were treated as 'neutral'. We will test a new approach of reclassifying these incorrect responses into the correct tags to see if this increases the scores of the different models. To do this, we will use the spaCy library, which specializes in natural language processing problems.

Using the different SpaCy models (SM, MD, and LG) and the \textbf{similarity()} function included, we created a function that takes as input an incorrect response and the tag list and returns as output the tag predicted with the highest semantic similarity to the incorrect response. To better observe the differences between using this approach and the approach without the use of dictionaries, the Rable \ref{tab:comparaison_resultats_3} will show for each model, in each case, the macro F1 scores, precision, recall, and accuracy obtained, with a precision of five decimal places.

\begin{table}[htb]
\centering
\setlength{\tabcolsep}{3pt}
\resizebox{\linewidth}{!}{%
\begin{tabular}{|| l | l | l | l | l | l | l | l ||} 
    \toprule
    Model name & Result type & Dictionary size & \textbf{Model macro F1 score (\%)} & \textbf{Precision (\%)} & \textbf{Recall (\%)} & \textbf{Accuracy (\%)} \\
    \hline
\midrule
    GPT-3.5-Turbo &  &  &  &  &  & \\ 
     & Original & N/A & 28.97477 & 31.96756 & 36.73094 & 23.71250 \\
     & With Dictionary & SM & 28.47332 & 31.96914 & 36.70862 & 22.77500  \\ 
     &  & MD & 28.66473 & 31.15335 & 37.59664 & 22.93750 \\ 
     &  & LG & 28.55063 & 30.71852 & 37.66528 & 22.97500 \\ \hline
     GPT-4o &  &  &  &  &  & \\ 
     & Original & N/A & 30.94547 & 32.48368 & 38.20576 & 24.10000 \\
     & With Dictionary & SM & 30.64027 & 31.99912 & 38.18851 & 23.28750 \\ 
     &  & MD & 30.64368 & 32.50769 & 38.50577 & 23.31250 \\ 
     &  & LG & 30.65712 & 32.25237 & 38.63715 & 23.37500 \\ \hline
     Gemini-1.5 &  &  &  &  &  & \\ 
     & Original & N/A & 26.74038 & 33.01710 & 31.71704 & 20.48750 \\
     & With Dictionary & SM & 26.65361 & 32.36802 & 31.74466 & 20.25000 \\ 
     &  & MD & 26.80581 & 32.84068 & 32.05650 & 20.21250 \\ 
     &  & LG & 26.84648 & 32.70287 & 32.13538 & 20.23750 \\ \hline
     Gemma-1.1-7b &  &  &  &  &  & \\ 
     & Original & N/A & 22.89399 & 33.10732 & 29.47269 & 22.66250 \\
     & With Dictionary & SM & 21.70670 & 32.57787 & 29.29456 & 18.15000 \\ 
     &  & MD & 22.91514 & 26.63251 & 31.50328 & 18.98750 \\ 
     &  & LG & 22.47977 & 27.28477 & 31.63337 & 18.98750 \\ \hline
     Gemma-2-9b &  &  &  &  &  & \\ 
     & Original & N/A & 24.20599 & 31.16082 & 31.66164 & 17.93750  \\
     & With Dictionary & SM & 23.95058 & 30.88174 & 31.65361 & 17.27500 \\ 
     &  & MD & 24.21847 & 30.79365 & 32.42775 & 17.32500 \\ 
     &  & LG & 24.37492 & 31.02687 & 32.64281 & 17.36250 \\ \hline
     Llama-2-7b &  &  &  &  &  & \\ 
     & Original & N/A & 20.24248 & 36.05048 & 21.86551 & 10.48750 \\
     & With Dictionary & SM & 19.60433 & 35.70024 & 21.96700 & 9.42500 \\ 
     &  & MD & 20.72441 & 28.03009 & 24.03709 & 9.33750 \\ 
     &  & LG & 20.60805 & 27.94807 & 24.37379 & 9.33750 \\ \hline
     Llama-3-8b &  &  &  &  &  & \\ 
     & Original & N/A & 20.59527 & 30.72653 & 27.36376 & 16.15000 \\
     & With Dictionary & SM & 20.12037 & 29.84229 & 27.33743 & 14.61250 \\ 
     &  & MD & 20.50476 & 27.82015 & 28.13073 & 14.85000 \\ 
     &  & LG & 20.31532 & 26.90405 & 28.27545 & 14.90000 \\ \hline
     Llama-3-70b &  &  &  &  &  & \\ 
     & Original & N/A & 27.20091 & 34.85671 & 33.62490 & 22.61250 \\
     & With Dictionary & SM & 26.96111 & 34.65644 & 33.65964 & 22.07500 \\ 
     &  & MD & 27.22124 & 34.01447 & 34.32349 & 22.16250 \\ 
     &  & LG & 27.25740 & 33.97677 & 34.41948 & 22.18750 \\ \hline
     Phi-3-4k &  &  &  &  &  & \\ 
     & Original & N/A & 25.23149 & 26.76328 & 37.07785 & 12.35000 \\
     & With Dictionary & SM & 24.79894 & 26.96175 & 37.13824 & 11.93750 \\ 
     &  & MD & 25.25119 & 27.53842 & 37.91341 & 12.05000 \\ 
     &  & LG & 26.01663 & 26.75966 & 38.70548 & 12.13750 \\ \hline
     Mistral-7b &  &  &  &  &  & \\ 
     & Original & N/A & 25.14239 & 28.64566 & 33.75372 & 18.97500 \\
     & With Dictionary & SM & 24.14168 & 28.36949 & 33.52505 & 16.41250 \\ 
     &  & MD & 24.69657 & 26.09023 & 35.59150 & 16.68750 \\ 
     &  & LG & 24.82202 & 25.76153 & 35.85328 & 16.75000 \\ \hline
     Mixtral-8x7b &  &  &  &  &  & \\ 
     & Original & N/A & 23.81649 & 28.67170 & 29.64733 & 19.63750 \\
     & With Dictionary & SM & 22.65727 & 27.31309 & 29.37475 & 15.43750 \\ 
     &  & MD & 16.22614 & 12.88130 & 35.40475 & 14.76250 \\ 
     &  & LG & 15.81165 & 12.31099 & 35.81679 & 14.77500\\ \hline
\bottomrule
\end{tabular}%
}
\caption{Comparison of the performance of different language models based on the use of dictionaries of different sizes}
\label{tab:comparaison_resultats_3}
\end{table}

Analysis of the results presented in Table \ref{tab:comparaison_resultats_3} shows that using dictionaries to reclassify incorrect responses has a variable impact on the performance of the different language models. Integrating dictionaries generally reduces macro F1 score and precision but improves recall. This drop in macro F1 score can be explained by invalid responses no longer being classified under the Neutral tag, further reducing the number of responses in this category. Large language models (LLMs) often have difficulty giving Neutral as an answer, so some of the wrong answers are counted as true Neutral positives.

For example, while improving recall, the GPT models see a notable decrease in precision when a dictionary is used. Similarly, the Mistral and Phi-3 models show similar trends, where the improvement in recall does not compensate for the loss in precision and macro F1 score. These observations confirm that the dictionary-based approach to correcting incorrect responses is not optimal for improving the overall performance of language models in emotion detection.

This method will not be used in the future, as automatic synonym search is an open problem that does not yield satisfactory results. The disparity in scores between different model sizes highlights the limitations of this approach, with performance varying significantly between small, medium, and large models.

\section{Limitations}\label{limits} 

While this study provides insights into LLMs' emotion detection capabilities, its primary reliance on the GoEmotions dataset displays limitations concerning generalizability. To further validate our findings, future research should explore their validity across datasets with diverse structures and assess model robustness against varying annotation schemes and cultural contexts. Incorporating benchmarks from multiple datasets could further validate our conclusions.

\section{Conclusion and Future Work}\label{conclusion}

In this work, we investigated the capabilities of LLMs in detecting and understanding human emotions through text, aiming to improve human-computer interaction by making AI technologies more responsive to emotional nuances. While we focused on statistical approaches using the GoEmotions dataset, we acknowledge that evaluating multiple datasets would strengthen the generality of our findings. Although ChatGPT and other LLMs demonstrate advanced text generation capabilities, their performances in emotion detection remain inferior to specialized models. However, applying prompt engineering techniques brought significant improvements, highlighting the importance of subtle guidance in eliciting more accurate responses. While LLMs may not surpass specialized classifiers like BERT in emotion detection tasks, the insights from this comparative study provide a valuable foundation for refining their performance.

Looking ahead, future efforts include introducing a new evaluation metric that accounts for semantic proximity between predicted and true emotions, rewarding near-correct predictions, and penalizing distant ones. Constructing a dedicated dialogue corpus would also allow more precise testing of a model's adaptability to linguistic and emotional nuances. Furthermore, future work will incorporate rigorous statistical validation to ensure that observed performance differences between models are statistically significant and not due to random chance

In conclusion, our research highlights the strengths and weaknesses of LLMs in emotion detection. Continuing this work could contribute to the evolution of artificial intelligence technologies, leading to a better understanding and a more empathetic response to human emotions.

\begin{acks}
The authors gratefully acknowledge the financial support provided by the European fond for regional development FEDER through the IA-EMOTIONS project. 
\end{acks}

\bibliographystyle{ACM-Reference-Format}
\bibliography{sample-base}


\begin{thebibliography}{41}


\ifx \showCODEN    \undefined \def \showCODEN     #1{\unskip}     \fi
\ifx \showDOI      \undefined \def \showDOI       #1{#1}\fi
\ifx \showISBNx    \undefined \def \showISBNx     #1{\unskip}     \fi
\ifx \showISBNxiii \undefined \def \showISBNxiii  #1{\unskip}     \fi
\ifx \showISSN     \undefined \def \showISSN      #1{\unskip}     \fi
\ifx \showLCCN     \undefined \def \showLCCN      #1{\unskip}     \fi
\ifx \shownote     \undefined \def \shownote      #1{#1}          \fi
\ifx \showarticletitle \undefined \def \showarticletitle #1{#1}   \fi
\ifx \showURL      \undefined \def \showURL       {\relax}        \fi
\providecommand\bibfield[2]{#2}
\providecommand\bibinfo[2]{#2}
\providecommand\natexlab[1]{#1}
\providecommand\showeprint[2][]{arXiv:#2}

\bibitem[noa({[n.\,d.]})]%
        {noauthor_monologgbert-base-cased-goemotions-original_nodate}
 \bibinfo{year}{[n.\,d.]}\natexlab{}.
\newblock \bibinfo{title}{Monologg/Bert-Base-Cased-Goemotions-Original · {{Hugging Face}}}.
\newblock
\newblock
\urldef\tempurl%
\url{https://huggingface.co/monologg/bert-base-cased-goemotions-original}
\showURL{%
\tempurl}


\bibitem[Lov(2024)]%
        {LovheimCubeEmotions2024}
 \bibinfo{year}{2024}\natexlab{}.
\newblock \showarticletitle{L{\"o}vheim {{Cube}} of {{Emotions}}}.
\newblock \bibinfo{journal}{\emph{Wikipedia}} (\bibinfo{date}{July} \bibinfo{year}{2024}).
\newblock


\bibitem[Abdin et~al\mbox{.}(2024)]%
        {abdin2024phi}
\bibfield{author}{\bibinfo{person}{Marah Abdin}, \bibinfo{person}{Jyoti Aneja}, \bibinfo{person}{Hany Awadalla}, \bibinfo{person}{Ahmed Awadallah}, \bibinfo{person}{Ammar~Ahmad Awan}, \bibinfo{person}{Nguyen Bach}, \bibinfo{person}{Amit Bahree}, \bibinfo{person}{Arash Bakhtiari}, \bibinfo{person}{Jianmin Bao}, \bibinfo{person}{Harkirat Behl}, {et~al\mbox{.}}} \bibinfo{year}{2024}\natexlab{}.
\newblock \showarticletitle{Phi-3 technical report: A highly capable language model locally on your phone}.
\newblock \bibinfo{journal}{\emph{arXiv preprint arXiv:2404.14219}} (\bibinfo{year}{2024}).
\newblock


\bibitem[Bostan et~al\mbox{.}({[n.\,d.]})]%
        {bostan_goodnewseveryone_2020}
\bibfield{author}{\bibinfo{person}{Laura Ana~Maria Bostan}, \bibinfo{person}{Evgeny Kim}, {and} \bibinfo{person}{Roman Klinger}.} \bibinfo{year}{[n.\,d.]}\natexlab{}.
\newblock \showarticletitle{{{GoodNewsEveryone}}: {{A Corpus}} of {{News Headlines Annotated}} with {{Emotions}}, {{Semantic Roles}}, and {{Reader Perception}}}. In \bibinfo{booktitle}{\emph{Proceedings of the {{Twelfth Language Resources}} and {{Evaluation Conference}}}} (Marseille, France, 2020-05), \bibfield{editor}{\bibinfo{person}{Nicoletta Calzolari}, \bibinfo{person}{Frédéric Béchet}, \bibinfo{person}{Philippe Blache}, \bibinfo{person}{Khalid Choukri}, \bibinfo{person}{Christopher Cieri}, \bibinfo{person}{Thierry Declerck}, \bibinfo{person}{Sara Goggi}, \bibinfo{person}{Hitoshi Isahara}, \bibinfo{person}{Bente Maegaard}, \bibinfo{person}{Joseph Mariani}, \bibinfo{person}{Hélène Mazo}, \bibinfo{person}{Asuncion Moreno}, \bibinfo{person}{Jan Odijk}, {and} \bibinfo{person}{Stelios Piperidis}} (Eds.). \bibinfo{publisher}{European Language Resources Association}, \bibinfo{pages}{1554--1566}.
\newblock
\showISBNx{979-10-95546-34-4}
\urldef\tempurl%
\url{https://aclanthology.org/2020.lrec-1.194}
\showURL{%
\tempurl}


\bibitem[Bota et~al\mbox{.}(2019)]%
        {article}
\bibfield{author}{\bibinfo{person}{Patricia Bota}, \bibinfo{person}{Chen Wang}, \bibinfo{person}{Ana Fred}, {and} \bibinfo{person}{Hugo Plácido~da Silva}.} \bibinfo{year}{2019}\natexlab{}.
\newblock \showarticletitle{A Review, Current Challenges, and Future Possibilities on Emotion Recognition Using Machine Learning and Physiological Signals}.
\newblock \bibinfo{journal}{\emph{IEEE Access}}  \bibinfo{volume}{PP} (\bibinfo{date}{09} \bibinfo{year}{2019}), \bibinfo{pages}{1--1}.
\newblock
\urldef\tempurl%
\url{https://doi.org/10.1109/ACCESS.2019.2944001}
\showDOI{\tempurl}


\bibitem[Bouazizi and Ohtsuki({[n.\,d.]})]%
        {bouazizi_sentiment_2016}
\bibfield{author}{\bibinfo{person}{Mondher Bouazizi} {and} \bibinfo{person}{Tomoaki Ohtsuki}.} \bibinfo{year}{[n.\,d.]}\natexlab{}.
\newblock \showarticletitle{Sentiment Analysis: {{From}} Binary to Multi-Class Classification: {{A}} Pattern-Based Approach for Multi-Class Sentiment Analysis in {{Twitter}}}. In \bibinfo{booktitle}{\emph{2016 {{IEEE International Conference}} on {{Communications}} ({{ICC}})}} (Kuala Lumpur, Malaysia, 2016-05). \bibinfo{publisher}{IEEE}, \bibinfo{pages}{1--6}.
\newblock
\showISBNx{978-1-4799-6664-6}
\urldef\tempurl%
\url{https://doi.org/10.1109/ICC.2016.7511392}
\showDOI{\tempurl}


\bibitem[Chernyavskiy et~al\mbox{.}({[n.\,d.]})]%
        {TransformersEndHistoryNLP_chernyavskiy_etal}
\bibfield{author}{\bibinfo{person}{Anton Chernyavskiy}, \bibinfo{person}{Dmitry Ilvovsky}, {and} \bibinfo{person}{Preslav Nakov}.} \bibinfo{year}{[n.\,d.]}\natexlab{}.
\newblock \bibinfo{booktitle}{\emph{Transformers: "{{The End}} of {{History}}" for {{NLP}}?}}
\newblock
\urldef\tempurl%
\url{https://doi.org/10.48550/arXiv.2105.00813}
\showDOI{\tempurl}
\showeprint[arXiv]{2105.00813}


\bibitem[Demszky et~al\mbox{.}({[n.\,d.]})]%
        {site_demszky_goemotions_2020}
\bibfield{author}{\bibinfo{person}{Dorottya Demszky}, \bibinfo{person}{Dana Movshovitz-Attias}, \bibinfo{person}{Jeongwoo Ko}, \bibinfo{person}{Alan Cowen}, \bibinfo{person}{Gaurav Nemade}, {and} \bibinfo{person}{Sujith Ravi}.} \bibinfo{year}{[n.\,d.]}\natexlab{}.
\newblock \showarticletitle{{{GoEmotions}}: {{A Dataset}} of {{Fine-Grained Emotions}}}. In \bibinfo{booktitle}{\emph{Proceedings of the 58th {{Annual Meeting}} of the {{Association}} for {{Computational Linguistics}}}} (Online, 2020-07), \bibfield{editor}{\bibinfo{person}{Dan Jurafsky}, \bibinfo{person}{Joyce Chai}, \bibinfo{person}{Natalie Schluter}, {and} \bibinfo{person}{Joel Tetreault}} (Eds.). \bibinfo{publisher}{Association for Computational Linguistics}.
\newblock
\urldef\tempurl%
\url{https://research.google/blog/goemotions-a-dataset-for-fine-grained-emotion-classification/}
\showURL{%
\tempurl}


\bibitem[Demszky et~al\mbox{.}(2020)]%
        {demszky_2020_goemotions}
\bibfield{author}{\bibinfo{person}{Dorottya Demszky}, \bibinfo{person}{Dana Movshovitz-Attias}, \bibinfo{person}{Jeongwoo Ko}, \bibinfo{person}{Alan Cowen}, \bibinfo{person}{Gaurav Nemade}, {and} \bibinfo{person}{Sujith Ravi}.} \bibinfo{year}{2020}\natexlab{}.
\newblock \showarticletitle{GoEmotions: A dataset of fine-grained emotions}.
\newblock \bibinfo{journal}{\emph{arXiv preprint arXiv:2005.00547}} (\bibinfo{year}{2020}).
\newblock


\bibitem[Devlin et~al\mbox{.}({[n.\,d.]})]%
        {DBLP:journals/corr/abs-1810-04805}
\bibfield{author}{\bibinfo{person}{Jacob Devlin}, \bibinfo{person}{Ming-Wei Chang}, \bibinfo{person}{Kenton Lee}, {and} \bibinfo{person}{Kristina Toutanova}.} \bibinfo{year}{[n.\,d.]}\natexlab{}.
\newblock \showarticletitle{{{BERT}}: Pre-Training of Deep Bidirectional Transformers for Language Understanding}.
\newblock   \bibinfo{volume}{abs/1810.04805} (\bibinfo{year}{[n.\,d.]}).
\newblock
\showeprint[arXiv]{1810.04805}
\urldef\tempurl%
\url{http://arxiv.org/abs/1810.04805}
\showURL{%
\tempurl}


\bibitem[Ekman({[n.\,d.]})]%
        {ekman_argument_1992}
\bibfield{author}{\bibinfo{person}{Paul Ekman}.} \bibinfo{year}{[n.\,d.]}\natexlab{}.
\newblock \showarticletitle{An Argument for Basic Emotions}.
\newblock  \bibinfo{volume}{6}, \bibinfo{number}{3-4} (\bibinfo{year}{[n.\,d.]}), \bibinfo{pages}{169--200}.
\newblock
\showISSN{0269-9931, 1464-0600}
\urldef\tempurl%
\url{https://doi.org/10.1080/02699939208411068}
\showDOI{\tempurl}


\bibitem[Ekman and Keltner(1970)]%
        {ekman1970universal}
\bibfield{author}{\bibinfo{person}{Paul Ekman} {and} \bibinfo{person}{Dacher Keltner}.} \bibinfo{year}{1970}\natexlab{}.
\newblock \showarticletitle{Universal facial expressions of emotion}.
\newblock \bibinfo{journal}{\emph{California mental health research digest}} \bibinfo{volume}{8}, \bibinfo{number}{4} (\bibinfo{year}{1970}), \bibinfo{pages}{151--158}.
\newblock


\bibitem[family=P S and G~S({[n.\,d.]})]%
        {p_s_emotion_2017}
\bibfield{author}{\bibinfo{person}{given-i=SREEJA family=P S, given=SREEJA} {and} \bibinfo{person}{Mahalakshmi G~S}.} \bibinfo{year}{[n.\,d.]}\natexlab{}.
\newblock \showarticletitle{Emotion {{Models}}: {{A Review}}}.
\newblock   \bibinfo{volume}{10} (\bibinfo{year}{[n.\,d.]}), \bibinfo{pages}{651--657}.
\newblock


\bibitem[Ferscha(2016)]%
        {ferscha2016research}
\bibfield{author}{\bibinfo{person}{Alois Ferscha}.} \bibinfo{year}{2016}\natexlab{}.
\newblock \showarticletitle{A research agenda for human computer confluence}.
\newblock \bibinfo{journal}{\emph{Human Computer Confluence Transforming Human Experience Through Symbiotic Technologies}} (\bibinfo{year}{2016}), \bibinfo{pages}{7--17}.
\newblock


\bibitem[Fox({[n.\,d.]})]%
        {Fox_Emotion_Science}
\bibfield{author}{\bibinfo{person}{Elaine Fox}.} \bibinfo{year}{[n.\,d.]}\natexlab{}.
\newblock \bibinfo{booktitle}{\emph{Emotion Science: {{Cognitive}} and Neuroscientific Approaches to Understanding Human Emotions}}.
\newblock
\showISBNx{978-0-230-00518-1}
\urldef\tempurl%
\url{https://doi.org/10.1007/978-1-137-07946-6}
\showDOI{\tempurl}


\bibitem[Frank and Wilson(2020)]%
        {frank2020silvan}
\bibfield{author}{\bibinfo{person}{Adam~J Frank} {and} \bibinfo{person}{Elizabeth~A Wilson}.} \bibinfo{year}{2020}\natexlab{}.
\newblock \bibinfo{booktitle}{\emph{A Silvan Tomkins handbook: Foundations for affect theory}}.
\newblock \bibinfo{publisher}{U of Minnesota Press}.
\newblock


\bibitem[Gillioz et~al\mbox{.}({[n.\,d.]})]%
        {OverviewTransformerbasedModelsNLPTasks_gillioz_etal}
\bibfield{author}{\bibinfo{person}{Anthony Gillioz}, \bibinfo{person}{Jacky Casas}, \bibinfo{person}{Elena Mugellini}, {and} \bibinfo{person}{Omar~Abou Khaled}.} \bibinfo{year}{[n.\,d.]}\natexlab{}.
\newblock \showarticletitle{Overview of the {{Transformer-based Models}} for {{NLP Tasks}}}. \bibinfo{pages}{179--183}.
\newblock
\urldef\tempurl%
\url{https://doi.org/10.15439/2020F20}
\showDOI{\tempurl}


\bibitem[Jiang et~al\mbox{.}(2023)]%
        {jiang2023mistral}
\bibfield{author}{\bibinfo{person}{Albert~Q Jiang}, \bibinfo{person}{Alexandre Sablayrolles}, \bibinfo{person}{Arthur Mensch}, \bibinfo{person}{Chris Bamford}, \bibinfo{person}{Devendra~Singh Chaplot}, \bibinfo{person}{Diego de~las Casas}, \bibinfo{person}{Florian Bressand}, \bibinfo{person}{Gianna Lengyel}, \bibinfo{person}{Guillaume Lample}, \bibinfo{person}{Lucile Saulnier}, {et~al\mbox{.}}} \bibinfo{year}{2023}\natexlab{}.
\newblock \showarticletitle{Mistral 7B}.
\newblock \bibinfo{journal}{\emph{arXiv preprint arXiv:2310.06825}} (\bibinfo{year}{2023}).
\newblock


\bibitem[Jiang et~al\mbox{.}(2024)]%
        {jiang2024mixtral}
\bibfield{author}{\bibinfo{person}{Albert~Q Jiang}, \bibinfo{person}{Alexandre Sablayrolles}, \bibinfo{person}{Antoine Roux}, \bibinfo{person}{Arthur Mensch}, \bibinfo{person}{Blanche Savary}, \bibinfo{person}{Chris Bamford}, \bibinfo{person}{Devendra~Singh Chaplot}, \bibinfo{person}{Diego de~las Casas}, \bibinfo{person}{Emma~Bou Hanna}, \bibinfo{person}{Florian Bressand}, {et~al\mbox{.}}} \bibinfo{year}{2024}\natexlab{}.
\newblock \showarticletitle{Mixtral of experts}.
\newblock \bibinfo{journal}{\emph{arXiv preprint arXiv:2401.04088}} (\bibinfo{year}{2024}).
\newblock


\bibitem[Jonnala({[n.\,d.]})]%
        {jonnala2024large}
\bibfield{author}{\bibinfo{person}{Alekya Jonnala}.} \bibinfo{year}{[n.\,d.]}\natexlab{}.
\newblock \showarticletitle{How {{Large Language}} Models ({{LLM}}) Help Enterprises Enhance Customer Experiences}.
\newblock  \bibinfo{volume}{13}, \bibinfo{number}{11} (\bibinfo{year}{[n.\,d.]}).
\newblock


\bibitem[Kajiwara et~al\mbox{.}({[n.\,d.]})]%
        {kajiwara_wrime_2021}
\bibfield{author}{\bibinfo{person}{Tomoyuki Kajiwara}, \bibinfo{person}{Chenhui Chu}, \bibinfo{person}{Noriko Takemura}, \bibinfo{person}{Yuta Nakashima}, {and} \bibinfo{person}{Hajime Nagahara}.} \bibinfo{year}{[n.\,d.]}\natexlab{}.
\newblock \showarticletitle{{{WRIME}}: {{A New Dataset}} for {{Emotional Intensity Estimation}} with {{Subjective}} and {{Objective Annotations}}}. In \bibinfo{booktitle}{\emph{Proceedings of the 2021 {{Conference}} of the {{North American Chapter}} of the {{Association}} for {{Computational Linguistics}}: {{Human Language Technologies}}}} (Online, 2021-06), \bibfield{editor}{\bibinfo{person}{Kristina Toutanova}, \bibinfo{person}{Anna Rumshisky}, \bibinfo{person}{Luke Zettlemoyer}, \bibinfo{person}{Dilek Hakkani-Tur}, \bibinfo{person}{Iz~Beltagy}, \bibinfo{person}{Steven Bethard}, \bibinfo{person}{Ryan Cotterell}, \bibinfo{person}{Tanmoy Chakraborty}, {and} \bibinfo{person}{Yichao Zhou}} (Eds.). \bibinfo{publisher}{Association for Computational Linguistics}, \bibinfo{pages}{2095--2104}.
\newblock
\urldef\tempurl%
\url{https://doi.org/10.18653/v1/2021.naacl-main.169}
\showDOI{\tempurl}


\bibitem[Kocoń et~al\mbox{.}({[n.\,d.]})]%
        {kocon_chatgpt_2023}
\bibfield{author}{\bibinfo{person}{Jan Kocoń}, \bibinfo{person}{Igor Cichecki}, \bibinfo{person}{Oliwier Kaszyca}, \bibinfo{person}{Mateusz Kochanek}, \bibinfo{person}{Dominika Szydło}, \bibinfo{person}{Joanna Baran}, \bibinfo{person}{Julita Bielaniewicz}, \bibinfo{person}{Marcin Gruza}, \bibinfo{person}{Arkadiusz Janz}, \bibinfo{person}{Kamil Kanclerz}, \bibinfo{person}{Anna Kocoń}, \bibinfo{person}{Bartłomiej Koptyra}, \bibinfo{person}{Wiktoria Mieleszczenko-Kowszewicz}, \bibinfo{person}{Piotr Miłkowski}, \bibinfo{person}{Marcin Oleksy}, \bibinfo{person}{Maciej Piasecki}, \bibinfo{person}{Łukasz Radliński}, \bibinfo{person}{Konrad Wojtasik}, \bibinfo{person}{Stanisław Woźniak}, {and} \bibinfo{person}{Przemysław Kazienko}.} \bibinfo{year}{[n.\,d.]}\natexlab{}.
\newblock \showarticletitle{{{ChatGPT}}: {{Jack}} of All Trades, Master of None}.
\newblock   \bibinfo{volume}{99} (\bibinfo{year}{[n.\,d.]}), \bibinfo{pages}{101861}.
\newblock
\showISSN{1566-2535}
\urldef\tempurl%
\url{https://doi.org/10.1016/j.inffus.2023.101861}
\showDOI{\tempurl}


\bibitem[Koutini et~al\mbox{.}({[n.\,d.]})]%
        {koutini_efficient_2022}
\bibfield{author}{\bibinfo{person}{Khaled Koutini}, \bibinfo{person}{Jan Schlüter}, \bibinfo{person}{Hamid Eghbal-zadeh}, {and} \bibinfo{person}{Gerhard Widmer}.} \bibinfo{year}{[n.\,d.]}\natexlab{}.
\newblock \showarticletitle{Efficient {{Training}} of {{Audio Transformers}} with {{Patchout}}}. In \bibinfo{booktitle}{\emph{Interspeech 2022}} (2022-09). \bibinfo{pages}{2753--2757}.
\newblock
\urldef\tempurl%
\url{https://doi.org/10.21437/Interspeech.2022-227}
\showDOI{\tempurl}


\bibitem[Lövheim({[n.\,d.]})]%
        {lovheim_new_2011}
\bibfield{author}{\bibinfo{person}{Hugo Lövheim}.} \bibinfo{year}{[n.\,d.]}\natexlab{}.
\newblock \showarticletitle{A New Three-Dimensional Model for Emotions and Monoamine Neurotransmitters}.
\newblock   \bibinfo{volume}{78} (\bibinfo{year}{[n.\,d.]}), \bibinfo{pages}{341--8}.
\newblock
\urldef\tempurl%
\url{https://doi.org/10.1016/j.mehy.2011.11.016}
\showDOI{\tempurl}


\bibitem[Obrenovic et~al\mbox{.}({[n.\,d.]})]%
        {obrenovic_generative_HCI_2024}
\bibfield{author}{\bibinfo{person}{Bojan Obrenovic}, \bibinfo{person}{Xiao Gu}, \bibinfo{person}{Guoyu Wang}, \bibinfo{person}{Danijela Godinić}, {and} \bibinfo{person}{Ilimdorjon Jakhongirov}.} \bibinfo{year}{[n.\,d.]}\natexlab{}.
\newblock \showarticletitle{Generative {{AI}} and Human-Robot Interaction: Implications and Future Agenda for Business, Society and Ethics}.
\newblock  (\bibinfo{year}{[n.\,d.]}).
\newblock
\urldef\tempurl%
\url{https://doi.org/10.1007/s00146-024-01889-0}
\showDOI{\tempurl}


\bibitem[Plutchik({[n.\,d.]})]%
        {plutchik_emotions_1991}
\bibfield{author}{\bibinfo{person}{Robert Plutchik}.} \bibinfo{year}{[n.\,d.]}\natexlab{}.
\newblock \bibinfo{booktitle}{\emph{The {{Emotions}}}}.
\newblock \bibinfo{publisher}{University Press of America}.
\newblock
\showISBNx{978-0-8191-8286-9}


\bibitem[Plutchik(1982)]%
        {plutchik_psychoevolutionary_1982}
\bibfield{author}{\bibinfo{person}{R Plutchik}.} \bibinfo{year}{1982}\natexlab{}.
\newblock \showarticletitle{A psycho evolutionary theory of emotions}.
\newblock \bibinfo{journal}{\emph{Social Science Information}} (\bibinfo{year}{1982}).
\newblock


\bibitem[Radford(2018)]%
        {radford2018improving}
\bibfield{author}{\bibinfo{person}{Alec Radford}.} \bibinfo{year}{2018}\natexlab{}.
\newblock \showarticletitle{Improving language understanding by generative pre-training}.
\newblock  (\bibinfo{year}{2018}).
\newblock


\bibitem[Rudolph et~al\mbox{.}(2023)]%
        {rudolph2023war}
\bibfield{author}{\bibinfo{person}{J{\"u}rgen Rudolph}, \bibinfo{person}{Shannon Tan}, {and} \bibinfo{person}{Samson Tan}.} \bibinfo{year}{2023}\natexlab{}.
\newblock \showarticletitle{War of the chatbots: Bard, Bing Chat, ChatGPT, Ernie and beyond. The new AI gold rush and its impact on higher education}.
\newblock \bibinfo{journal}{\emph{Journal of Applied Learning and Teaching}} \bibinfo{volume}{6}, \bibinfo{number}{1} (\bibinfo{year}{2023}), \bibinfo{pages}{364--389}.
\newblock


\bibitem[Saravia et~al\mbox{.}({[n.\,d.]})]%
        {saravia-etal-2018-carer}
\bibfield{author}{\bibinfo{person}{Elvis Saravia}, \bibinfo{person}{Hsien-Chi~Toby Liu}, \bibinfo{person}{Yen-Hao Huang}, \bibinfo{person}{Junlin Wu}, {and} \bibinfo{person}{Yi-Shin Chen}.} \bibinfo{year}{[n.\,d.]}\natexlab{}.
\newblock \showarticletitle{{{CARER}}: {{Contextualized}} Affect Representations for Emotion Recognition}. In \bibinfo{booktitle}{\emph{Proceedings of the 2018 Conference on Empirical Methods in Natural Language Processing}} (Brussels, Belgium, 0010/2018-11). \bibinfo{publisher}{Association for Computational Linguistics}, \bibinfo{pages}{3687--3697}.
\newblock
\urldef\tempurl%
\url{https://doi.org/10.18653/v1/D18-1404}
\showDOI{\tempurl}


\bibitem[Scherer and Wallbott({[n.\,d.]})]%
        {scherer_evidence_1994}
\bibfield{author}{\bibinfo{person}{Klaus~R. Scherer} {and} \bibinfo{person}{Harald~G. Wallbott}.} \bibinfo{year}{[n.\,d.]}\natexlab{}.
\newblock \showarticletitle{Evidence for Universality and Cultural Variation of Differential Emotion Response Patterning.}
\newblock  \bibinfo{volume}{66}, \bibinfo{number}{2} (\bibinfo{year}{[n.\,d.]}), \bibinfo{pages}{310--328}.
\newblock
\showISSN{1939-1315, 0022-3514}
\urldef\tempurl%
\url{https://doi.org/10.1037/0022-3514.66.2.310}
\showDOI{\tempurl}


\bibitem[Shao(2023)]%
        {shao2023empathetic}
\bibfield{author}{\bibinfo{person}{Ruosi Shao}.} \bibinfo{year}{2023}\natexlab{}.
\newblock \showarticletitle{An Empathetic AI for Mental Health Intervention: Conceptualizing and Examining Artificial Empathy}. In \bibinfo{booktitle}{\emph{Proceedings of the 2nd Empathy-Centric Design Workshop}}. \bibinfo{pages}{1--6}.
\newblock


\bibitem[Team et~al\mbox{.}(2024a)]%
        {team2024gemini}
\bibfield{author}{\bibinfo{person}{Gemini Team}, \bibinfo{person}{Petko Georgiev}, \bibinfo{person}{Ving~Ian Lei}, \bibinfo{person}{Ryan Burnell}, \bibinfo{person}{Libin Bai}, \bibinfo{person}{Anmol Gulati}, \bibinfo{person}{Garrett Tanzer}, \bibinfo{person}{Damien Vincent}, \bibinfo{person}{Zhufeng Pan}, \bibinfo{person}{Shibo Wang}, {et~al\mbox{.}}} \bibinfo{year}{2024}\natexlab{a}.
\newblock \showarticletitle{Gemini 1.5: Unlocking multimodal understanding across millions of tokens of context}.
\newblock \bibinfo{journal}{\emph{arXiv preprint arXiv:2403.05530}} (\bibinfo{year}{2024}).
\newblock


\bibitem[Team et~al\mbox{.}(2024b)]%
        {team2024gemma}
\bibfield{author}{\bibinfo{person}{Gemma Team}, \bibinfo{person}{Thomas Mesnard}, \bibinfo{person}{Cassidy Hardin}, \bibinfo{person}{Robert Dadashi}, \bibinfo{person}{Surya Bhupatiraju}, \bibinfo{person}{Shreya Pathak}, \bibinfo{person}{Laurent Sifre}, \bibinfo{person}{Morgane Rivi{\`e}re}, \bibinfo{person}{Mihir~Sanjay Kale}, \bibinfo{person}{Juliette Love}, {et~al\mbox{.}}} \bibinfo{year}{2024}\natexlab{b}.
\newblock \showarticletitle{Gemma: Open models based on gemini research and technology}.
\newblock \bibinfo{journal}{\emph{arXiv preprint arXiv:2403.08295}} (\bibinfo{year}{2024}).
\newblock


\bibitem[the~free encyclopedia({[n.\,d.]})]%
        {Plutchik_wheel_2024}
\bibfield{author}{\bibinfo{person}{From~Wikipedia the~free encyclopedia}.} \bibinfo{year}{[n.\,d.]}\natexlab{}.
\newblock \bibinfo{title}{Robert {{Plutchik}}}.
\newblock
\newblock
\urldef\tempurl%
\url{https://en.wikipedia.org/w/index.php?title=Robert_Plutchik&oldid=1240659436}
\showURL{%
\tempurl}


\bibitem[Tomkins({[n.\,d.]})]%
        {tomkins_affect_1962}
\bibfield{author}{\bibinfo{person}{Silvan Tomkins}.} \bibinfo{year}{[n.\,d.]}\natexlab{}.
\newblock \bibinfo{booktitle}{\emph{Affect {{Imagery Consciousness}}: {{Volume I}}: {{The Positive Affects}}}}.
\newblock \bibinfo{publisher}{Springer Publishing Company}.
\newblock
\showISBNx{978-0-8261-0442-7}


\bibitem[Touvron et~al\mbox{.}(2023)]%
        {touvron2023llama}
\bibfield{author}{\bibinfo{person}{Hugo Touvron}, \bibinfo{person}{Thibaut Lavril}, \bibinfo{person}{Gautier Izacard}, \bibinfo{person}{Xavier Martinet}, \bibinfo{person}{Marie-Anne Lachaux}, \bibinfo{person}{Timoth{\'e}e Lacroix}, \bibinfo{person}{Baptiste Rozi{\`e}re}, \bibinfo{person}{Naman Goyal}, \bibinfo{person}{Eric Hambro}, \bibinfo{person}{Faisal Azhar}, {et~al\mbox{.}}} \bibinfo{year}{2023}\natexlab{}.
\newblock \showarticletitle{Llama: Open and efficient foundation language models}.
\newblock \bibinfo{journal}{\emph{arXiv preprint arXiv:2302.13971}} (\bibinfo{year}{2023}).
\newblock


\bibitem[Vaswani(2017)]%
        {vaswani_2017_attention}
\bibfield{author}{\bibinfo{person}{A Vaswani}.} \bibinfo{year}{2017}\natexlab{}.
\newblock \showarticletitle{Attention is all you need}.
\newblock \bibinfo{journal}{\emph{Advances in Neural Information Processing Systems}} (\bibinfo{year}{2017}).
\newblock


\bibitem[Wermter et~al\mbox{.}({[n.\,d.]})]%
        {wermter_connectionist_1996}
\bibfield{author}{\bibinfo{person}{Stefan Wermter}, \bibinfo{person}{Ellen Riloff}, {and} \bibinfo{person}{Gabriele Scheler}.} \bibinfo{year}{[n.\,d.]}\natexlab{}.
\newblock \bibinfo{booktitle}{\emph{Connectionist, {{Statistical}} and {{Symbolic Approaches}} to {{Learning}} for {{Natural Language Processing}}}}.
\newblock \bibinfo{publisher}{Springer Science \& Business Media}.
\newblock
\showISBNx{978-3-540-60925-4}


\bibitem[Xu et~al\mbox{.}({[n.\,d.]})]%
        {xu2023can}
\bibfield{author}{\bibinfo{person}{Zhenyu Xu}, \bibinfo{person}{Hailin Xu}, \bibinfo{person}{Zhouyang Lu}, \bibinfo{person}{Yingying Zhao}, \bibinfo{person}{Rui Zhu}, \bibinfo{person}{Yujiang Wang}, \bibinfo{person}{Mingzhi Dong}, \bibinfo{person}{Yuhu Chang}, \bibinfo{person}{Qin Lv}, \bibinfo{person}{Robert~P Dick}, {et~al\mbox{.}}} \bibinfo{year}{[n.\,d.]}\natexlab{}.
\newblock \bibinfo{title}{Can Large Language Models Be Good Companions? {{An LLM-based}} Eyewear System with Conversational Common Ground}.  (\bibinfo{year}{[n.\,d.]}).
\newblock
\showeprint[arXiv]{2311.18251}


\bibitem[Yadollahi et~al\mbox{.}({[n.\,d.]})]%
        {yadollahi_current_2018}
\bibfield{author}{\bibinfo{person}{Ali Yadollahi}, \bibinfo{person}{Ameneh~Gholipour Shahraki}, {and} \bibinfo{person}{Osmar~R. Zaiane}.} \bibinfo{year}{[n.\,d.]}\natexlab{}.
\newblock \showarticletitle{Current {{State}} of {{Text Sentiment Analysis}} from {{Opinion}} to {{Emotion Mining}}}.
\newblock  \bibinfo{volume}{50}, \bibinfo{number}{2} (\bibinfo{year}{[n.\,d.]}), \bibinfo{pages}{1--33}.
\newblock
\showISSN{0360-0300, 1557-7341}
\urldef\tempurl%
\url{https://doi.org/10.1145/3057270}
\showDOI{\tempurl}


\end{thebibliography}

\end{document}